\documentclass[10pt,twocolumn,letterpaper]{article}

\usepackage{iccv}
\usepackage{times}
\usepackage{epsfig}
\usepackage{graphicx}
\usepackage{amsmath}
\usepackage{amssymb}
\usepackage{color}
\usepackage{tabularx}
\usepackage{array}

\usepackage[pagebackref=true,breaklinks=true,letterpaper=true,colorlinks,bookmarks=false]{hyperref}

\iccvfinalcopy 

\def\iccvPaperID{1234}

\ificcvfinal\fi
\begin{document}

\title{Deep Learning for Single-View Instance Recognition}

\author{David Held, Sebastian Thrun, Silvio Savarese\\
Stanford University\\
{\tt\small \{davheld, thrun, ssilvio\}@cs.stanford.edu}
}

\maketitle

\begin{abstract}
Deep learning methods have typically been trained on large datasets in which many training examples are available.  However, many real-world product datasets have only a small number of images available for each product.  We explore the use of deep learning methods for recognizing object instances when we have only a single training example per class.  We show that feedforward neural networks outperform state-of-the-art methods for recognizing objects from novel viewpoints even when trained from just a single image per object. To further improve our performance on this task, we propose to take advantage of a supplementary dataset in which we observe a separate set of objects from multiple viewpoints.  We introduce a new approach for training deep learning methods for instance recognition with limited training data, in which we use an auxiliary multi-view dataset to train our network to be robust to viewpoint changes.  We find that this approach leads to a more robust classifier for recognizing objects from 
novel viewpoints, outperforming previous state-of-the-art approaches including keypoint-matching, template-based techniques, and sparse coding.  
\end{abstract}

\section{Introduction}

There are many real-world scenarios in which we want to recognize an object instance from just a single training example.  For example, for many product databases available on Amazon, Safeway, or other websites, only a small number of images are available for each product.  Given a novel viewpoint of a product, can we robustly recognize the target object?  

Enabling a computer vision system to recognize objects from just one training example would enable a range of applications that could train on images from product databases.  For example, a kitchen perception system might need to recognize the grocery products in the kitchen.  Such a system would ideally be trained from a grocery product database, even if only one image of each product were available.  

We would also like to enable casual users to train a classifier to recognize an object after taking just a single picture of the target object.  
Such a method could be used to bootstrap a number of custom applications that require understanding how a user interacts with the objects in their environment.  Thus, we need computer vision methods that can robustly recognize objects after training from just a single image.

\begin{figure}[t]
\centering
\includegraphics[width=0.49\textwidth]{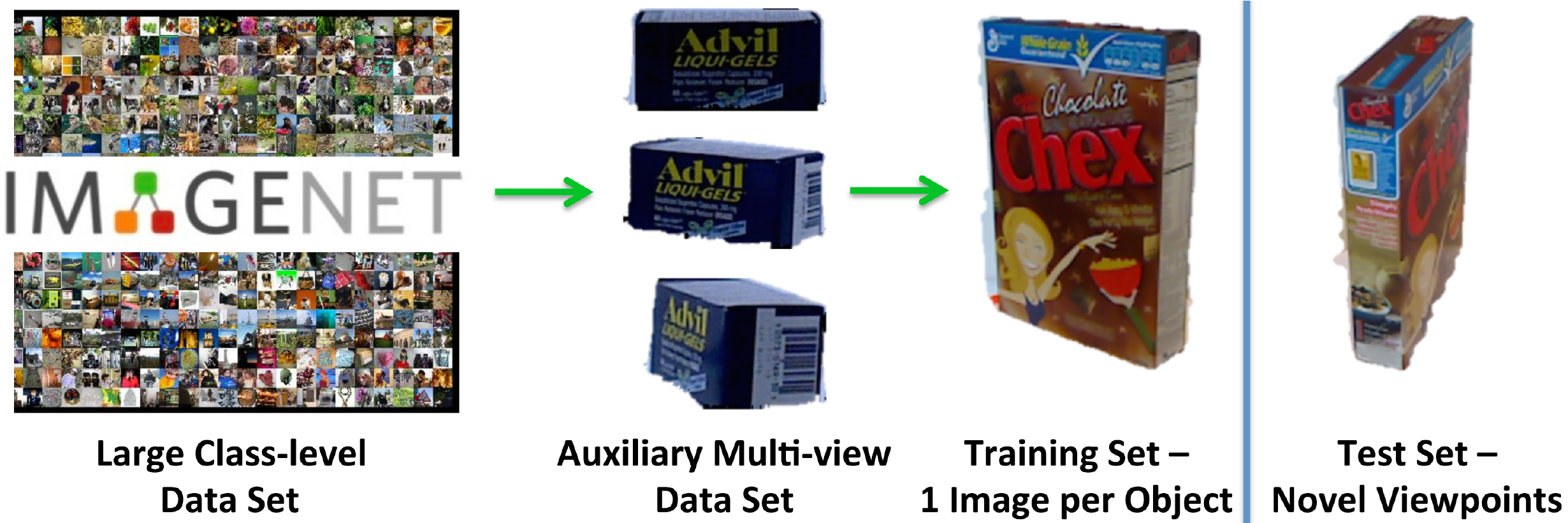} %
\caption{Given only a single image of an object, we want to recognize this object from novel viewpoints.  We perform a multi-stage training procedure, in which we first pre-train on a large class-level dataset, followed by an auxiliary multi-view dataset,  which trains our network to be robust to viewpoint changes.  Finally we train on the objects we wish to recognize from just a single image.}
\label{fig:pull}
\end{figure}

Traditionally, researchers have used feature-matching based approaches to recognize objects from a single example.  Unfortunately, because feature-matching approaches rely on being able to detect distinctive keypoints on an object, they often fail for textureless objects or for non-planar objects with large changes in viewpoint~\cite{moreels2007evaluation}.  
Machine learning methods, including deep learning, have been successfully applied to recognize objects at the class level~\cite{krizhevsky2012imagenet, girshick2014rich}, but they are not commonly used to recognize specific object instances, especially when only a single training image is available for each object.  

We introduce a new approach to training neural networks to recognize objects from just a single image, using a general-to-specific training procedure.  We initially use a large dataset to train our network to recognize general object classes.  We then train our network on a smaller dataset in which we observe objects from multiple viewpoints.  Finally, we train our network on a separate dataset in which only a single image is available for each object instance, as depicted in Figure~\ref{fig:pull}.  

By training our network in this general-to-specific manner, our network learns the invariances that it needs to perform the final task.  Our network initially learns general visual properties about the world.  It then learns generic object invariances, enabling the network to be robust to rotations and changes in viewpoint.  Finally, our network learns to recognize a specific set of objects from just a single training image per object.  

Using this novel multi-stage training procedure, our network learns to robustly recognize objects from new viewpoints.  To our knowledge, this is the first work that uses deep learning to recognize specific object instances from a single image.  We perform an extensive evaluation and show that multi-view pre-training outperforms previous state-of-the-art approaches for recognizing both textured and untextured objects from novel viewpoints.  

\section{Related Work}
Instance recognition has traditionally been achieved using either keypoint-based
methods~\cite{lowe2004distinctive, bay2006surf} or by matching local image
patches~\cite{ferrari2004simultaneous, rothganger20063d, lazebnik2003sparse}.
Keypoints can be filtered using different criteria~\cite{lowe2004distinctive}
and validated using RANSAC or Hough Voting to ensure geometric
consistency~\cite{fischler1981random}.  Although keypoint-based approaches have
shown some success for image recognition~\cite{chen2013residual}, such methods
are unreliable for recognizing untextured objects or non-planar objects when the
viewpoint is changed by more than 25 degrees~\cite{moreels2007evaluation}.  

Template matching has also been used for instance
recognition~\cite{huttenlocher1993comparing, olson1997automatic}.  Much work has
recently been done to make template matching scalable, efficient, and robust to
occlusions~\cite{hinterstoisser2012gradient, hsiao2012occlusion, damen2012real}.
 However, viewpoint invariance is usually achieved by recording many templates
during training from different viewing angles.  If only a small number of images
are available from each object during training, template matching methods will
not robustly detect the target object, as we will demonstrate.

Another approach that can be used for recognizing objects is to use machine
learning methods to train an object classifier~\cite{lazebnik2006beyond,
bo2013unsupervised}.  One example of such a classifier that has shown great
success in recent years is a convolutional neural
network~\cite{fukushima1980neocognitron, lecun1998gradient,
krizhevsky2012imagenet}.  However,  statistical methods such as neural networks
typically require many training examples to perform well.  For example, for the
ImageNet challenge, participants train their methods on 1.2 million training
examples~\cite{deng2009imagenet}.   Recently, some groups have successfully
trained their networks on just 5000 images across 20
classes~\cite{agrawal2014analyzing}, sometimes using domain-specific
fine-tuning~\cite{girshick2014rich}. We will test the performance of neural
networks when only 1 training example is available per class.  

One-shot learning has also been explored for classifying objects at the
category-level~\cite{fei2006one, tommasi2009more, salakhutdinov2010one} or
for recognizing handwritten characters~\cite{lake2011one, lake2013one}.  In 
constrast, we focus on recognizing object instances from novel viewpoints, and we compare
our approach to state of the art techniques for object instance recognition.

Our method makes use of a separate multi-view dataset to improve performance on
the task of instance recognition from a single training image.  Our idea of
using a supplemental multi-view dataset is related to previous efforts to
improve recognition performance by using a video sequence~\cite{mobahi2009deep,
becker1997learning}.
Another related effort is to use unlabeled video for
unsupervised feature learning~\cite{leistner2011improving, wiskott2002slow}. 
These methods typically enforce the consistency of features between subsequent
video frames.  We instead use multi-view objects in a classification setting to
improve our performance for recognizing single-view objects, and we do not treat
the multi-view dataset as a linear video sequence.

Some researchers have attempted to measure the invariance of deep networks to
rotations and other types of transformations~\cite{goodfellow2009measuring,
peng2014exploring, lenc2014understanding}.  These papers have focused on
measuring the rotational invariance for image patches or for general object
classes rather than object instances. 

The problem of adapting a class-level classifier for use in viewpoint-invariant instance recognition
is also related to the topic of domain adaptation~\cite{bergamo2010exploiting,
hoffman2013one, hoffman2014asymmetric, ganin2014unsupervised}. 
However, in most domain adaptation problems, 
the source dataset contains the same object labels as during inference. 
The goal of domain adaptation is to adapt a
classifier trained from a given “source” domain (e.g.~Amazon images) to
classify the same set of objects in the “target” domain (e.g.~Webcam images).  
On the other hand, for our task the classifier must learn to recognize novel object instances,
with no overlap between our large ``source" datasets and the final query images that we wish to recognize.
Further, we are interested in adapting class-level classifiers for viewpoint
invariant instance recognition, which has not been explored previously.

\section{Method}
\subsection{Problem Setup}
Suppose that we are given an image $x_i$ of an object instance that we want to recognize.  We assume that we have a ``single-view" database of $K_{S}$ different objects, and that the object in our image $x_i$ is one of the $K_{S}$ objects in our single-view database.  We also assume that each of the objects in our database has only one image taken of it.  Given that our image $x_i$ is likely to be taken from a novel viewpoint relative to the images in our database, how can we robustly identify the instance label for this object?

In order to robustly perform this task, we suppose that we also have a separate ``multi-view" set of $K_{M}$ objects for which we have recorded images from many viewpoints.   Because we have observed each of these separate objects from many viewing angles, we can use these images to teach our method to be invariant to viewpoint changes.  Then, given a novel viewpoint of an object from the single-view dataset, we can use this learned invariance to correctly recognize the target object.  

Note that the multi-view objects are chosen so that there is no overlap between the $K_M$ multi-view objects and the $K_S$ single-view objects.  Thus, any invariances that we learn from the multi-view dataset must be general to be able to transfer over to a new set of objects.  Our final goal is to identify an image $x_i$ as belonging to one of the $K_{S}$ single-view objects; the multi-view dataset is helpful only in teaching our method to be invariant to viewpoint changes.

\subsection{Multi-View Pre-Training}
\label{sec:joint_training}
We consider instance recognition as a classification problem, and we will explore the use of neural networks to perform this task.
Because neural networks represent a non-convex decision boundary, the initialization of the network is important.  One common approach for training a neural network with a limited amount of data is to initialize the network by pre-training on a larger dataset~\cite{girshick2014rich} (\eg ImageNet~\cite{deng2009imagenet}).  These initial weights are then fine-tuned using a smaller dataset for the relevant task.  This training procedure allows the network to find a better local optimum.

However, the ability to transfer information from the larger dataset to the smaller dataset, via network initialization, depends on the similarity between the datasets.  If the datasets are not very similar, then this initialization will be poor~\cite{yosinski2014transferable}.  As we will show, pre-training the network for class-level recognition (\eg using ImageNet) is not ideal for training these networks to be viewpoint invariant with respect to specific object instances.  

For the original ImageNet classification task, the goal of the network is to recognize 1000 different object classes.  Each class represents an object category, such as ``restaurant" or ``mask," and the appearance of objects within the class can vary dramatically; different restaurants can have a very different appearance. %
Because the network must recognize generic object classes, the computational effort of the network is spent attempting to handle all of the different aspects of intra-class variability.  
On the other hand, if our goal is to perform object instance recognition, then we can focus our network's computational effort on being robust to rotations, leading to better performance at this task.

We will show that, although pre-training our network on ImageNet provides a decent initialization for our network, we can obtain better performance through a multi-stage training procedure, as follows:
\begin{enumerate} \itemsep1pt \parskip0pt \parsep0pt
\item Train our network on a large class-level dataset.
\item Train our network on an instance-level dataset with many views per object instance.
\item Train our network to recognize a new set of object instances from a single image per object.
\end{enumerate}
This setup is illustrated in Figure~\ref{fig:pull}.  In more detail, we initially pre-train our network on a large class-level dataset, \eg ImageNet, which allows our network to learn general image statistics.  We then train our network on a smaller dataset in which we observe a set of objects from multiple viewpoints, and we learn to recognize these objects instances.  This stage allows our network to learn to be robust to changes in viewpoint.  Finally, we train our network on a separate dataset in which only a single image is available for each object.  We show that adding an intermediate multi-view pre-training step (step 2 above) gives better performance than pre-training only on a class-level dataset.  Adding multi-view pre-training increases the robustness of our network and enables us to recognize novel objects from new viewpoints.

We would also like to be able to recognize objects in real scenes against random
backgrounds.  To make our network robust to different backgrounds, during
multi-view pre-training (step 2) we synthetically place the objects against
random background scenes which do not contain any of the test objects.  Although
the single-view objects that we wish to recognize are placed against a fixed
background for training (in step 3), we will show that pre-training with separate multi-view
objects in step 2 against random backgrounds allows our method to learn to be
robust to new backgrounds.  

One can view our approach as an extension of data augmentation techniques for neural networks.  It is common when training neural networks to perform multiple image transformations on each training example to synthetically generate more training examples.  Common transformations include crops, horizontal flips, and lighting changes~\cite{krizhevsky2012imagenet}.

These data augmentation methods are an attempt to train the network to be robust to translations or changes in lighting.  However, it is more difficult to construct an image transformation that would simulate an out-of-plane rotation.  As an alternative, we propose multi-view pre-training, in which our intermediate stage involves classifying a separate set of objects when trained from multiple viewpoints.  Multi-view pre-training allows our network to learn new kinds of invariances, such as out-of-plane rotations, that would be hard to simulate using data augmentations. 

\subsection{Network Details}
\label{sec:network_details}

Our neural network uses the CaffeNet architecture~\cite{jia2014caffe}, which is very similar to the architecture proposed by Krizhevsky \etal~\cite{krizhevsky2012imagenet}.
The network is initially pre-trained on ImageNet~\cite{deng2009imagenet}.  We then fine-tune this network on the multi-view dataset as follows: we replace the final layer with a $K_{M}$ class classifier, and we fine-tune the weights to classify the $K_{M}$ multi-view objects.  We call this step ``multi-view pre-training" since we are training the network to recognize object instances given multiple views of each object.  
During multi-view pre-training, we hold the convolutional layers fixed and only fine-tune the fully-connected layers on top.  The number of layers that we fix was determined using a hold-out validation set.  

During multi-view pre-training, we use a learning rate of 0.001 for all layers except the final layer, which we set to a learning rate of 0.01. After 50,000 iterations, we reduce the learning rate by a factor of 10, and after 100,000 iterations we stop the multi-view training.  Other hyperparameters are taken from the default parameters for CaffeNet~\cite{jia2014caffe}, and we left them unchanged.  

Finally, we initialize the network using the learned weights from multi-view pre-training, and we fine-tune the network to classify the single-view objects.  To do this, we replace the final layer with a $K_{S}$ class classifier for the $K_{S}$ single-view objects. Each object in this dataset has only 1 training example from a single viewpoint.  We use the same parameters as before, except that the learning rates are reduced by a factor of 10, which was again determined using cross-validation on a hold-out set.  The final classifier is used to classify these $K_{S}$ objects from novel viewpoints.  We call a classifier trained in this manner a ``neural network with multi-view pre-training."  

\section{Results}
We perform a number of experiments to analyze the performance of different instance recognition methods.  In Sections~\ref{sec:dense_training_setup} through~\ref{sec:one_shot}, we use the RGB-D object dataset~\cite{lai2011large}, in which we recognize objects that are placed on a turntable and recorded from different viewpoints.  In this controlled setup, we can measure the object's angular difference between the training and test images, allowing us to compute how robust the different methods are to out-of-plane rotations.  We will later evaluate the methods on recognition in real-world scenes, as described below.

We evaluate the performance of different methods on this dataset under three conditions:
\begin{enumerate} \itemsep1pt \parskip0pt \parsep0pt
\item Training from many examples (Section \ref{sec:dense_training_setup})
\item Training from a variable number of examples (Section \ref{sec:sparse_training_setup})
\item Training from just a single example (Section \ref{sec:one_shot})
\end{enumerate}
In all three cases, we use the same test set, which is the instance recognition test set from~\cite{lai2011large}.  None of the methods that were evaluated use depth information except to segment out the target object. 

We vary the number of examples available during training to show how well each method generalizes with a limited number of training examples.   When multiple training examples are available, we compare the neural network-based approaches to other machine learning approaches.  When only one training example is available, we also evaluate keypoint-matching and other approaches that are designed to match pairs of images.  We find that neural networks have superior performance in all three cases, and we further show the advantage of multi-view pre-training in the case of training from just a single example.

Finally, in Section~\ref{sec:occlusion_handling}, we evaluate how robust the different classifiers are to handling occlusions and real backgrounds.  For this we use the RGB-D scenes dataset~\cite{lai2011large}, in which the objects from the previous test set are placed in a real-world scene.  Our task now is to recognize the object given the object's bounding box.  In an end-to-end system, the bounding box would be generated using one of the many methods that have been developed for this purpose~\cite{uijlings2013selective, alexe2012measuring, carreira2010constrained, krahenbuhl2014geodesic, cheng2014bing, manen2013prime}.  Although we can no longer compute the angular difference between training and test images in this less controlled setting, this experiment allows us to determine how robust the different methods are to recognition against a real background and under occlusions.  For this task, we use the same training set as before, \ie training from just a single example.  We also evaluate the performance 
as a function of the noise in the bounding box location and show that our method is robust to such variations.

\subsection{Instance recognition from many examples}
\label{sec:dense_training_setup}
We first evaluate our method using the RGB-D object dataset~\cite{lai2011large}, and we measure the performance when many training examples are available.  This dataset consists of 300 objects of different types and textures, ranging from apples to cereal boxes.  Given an image of one of these objects taken from a novel viewpoint, our task is to identify which of the 300 objects this image is taken from.  We treat this task as a 300-class classification problem, and we are thus able to apply tools from machine learning to perform this task.  Although the dataset consists of RGB images as well as depth, we only use the depth to obtain a segmentation of the target objects, both during training and at test time.  In Section~\ref{sec:occlusion_handling} we will explore the performance of the different methods when objects are placed in a real scene where a segmentation mask is not available.

We initially evaluate our method using the ``leave sequence out" training set up of~\cite{lai2011large}.  In this setup, we observe each object at a 30 degree and 60 degree elevation angle during training, and we observe the object at a 45 degree elevation angle at test time.  During training, the object is placed on a turn-table, and we observe the object from many views spaced 6 to 9 degrees apart in azimuth. 

The results for this setup when many training images are available can be seen in Table~\ref{table:dense_training_tab}.  We compare to the method of~\cite{lai2011large}, which combines a number of visual features, including dense SIFT~\cite{lowe2004distinctive}, texton histograms~\cite{leung2001representing}, and a color histogram.  The methods that learn feature descriptors, such as~\cite{blum2012learned} and~\cite{bo2013unsupervised} do significantly better, achieving accuracies between 90.4 and 92.1\%.  The results from~\cite{bo2013unsupervised} indicate that only small gains are achieved by adding depth information.

We evaluate the performance of a neural network pre-trained only with ImageNet (with no multi-view pre-training).  Using such a network, we are able to outperform all of these previous methods, obtaining an accuracy of 93.3\%.  When many training examples are available for each object, we can achieve high performance without multi-view pre-training.  However, in Section~\ref{sec:sparse_training_setup} we show that pre-training only on ImageNet has poor performance when the number of training examples per object is limited, thus motivating the use of multi-view pre-training for such cases. 

\begin{table}
\begin{tabular}{| l | c | }
\hline
Method & \% Accuracy \\
\hline
SIFT + Texton + Color Histogram \cite{lai2011large} & 60.7 \\
Convolutional k-means descriptor~\cite{blum2012learned} & 90.4 \\
HMP (RGB) \cite{bo2013unsupervised} &  92.1 \\
\textbf{Neural network (Ours)} & \textbf{93.3} \\
\hline
\end{tabular}
\caption{Training from many views: We first compare a neural network to previous methods for instance recognition when many views of each object instance are available during training.  The neural network that we evaluate here is pre-trained only on ImageNet, with no multi-view pre-training.}
\label{table:dense_training_tab}
\end{table}

\subsection{Varying the number of training images}
\label{sec:sparse_training_setup}
We note that the training setup from~\cite{lai2011large} is somewhat unrealistic.  It is rare that objects are placed on a turn-table and that someone records images at so many different angles and elevations during training.  In a typical product database, it is much more common to have only a few images taken of each object of interest.  Further, a casual user will want to be able to recognize an object after taking only a few pictures during training.

We therefore create a new training setup to test the performance of these methods in a more realistic scenario.  Each object is now viewed at training time at only a 30 degree elevation angle.  We also vary the number of azimuthal angles for which an object is observed in training from 69 viewpoints down to just a single viewpoint.  We evenly sample from the available training images for each object, starting from the first image.  We can thus use this setup to determine how the performance of different methods are affected by the number of training examples.

\begin{figure}[htb]
\centering
\includegraphics[width=0.33\textwidth]{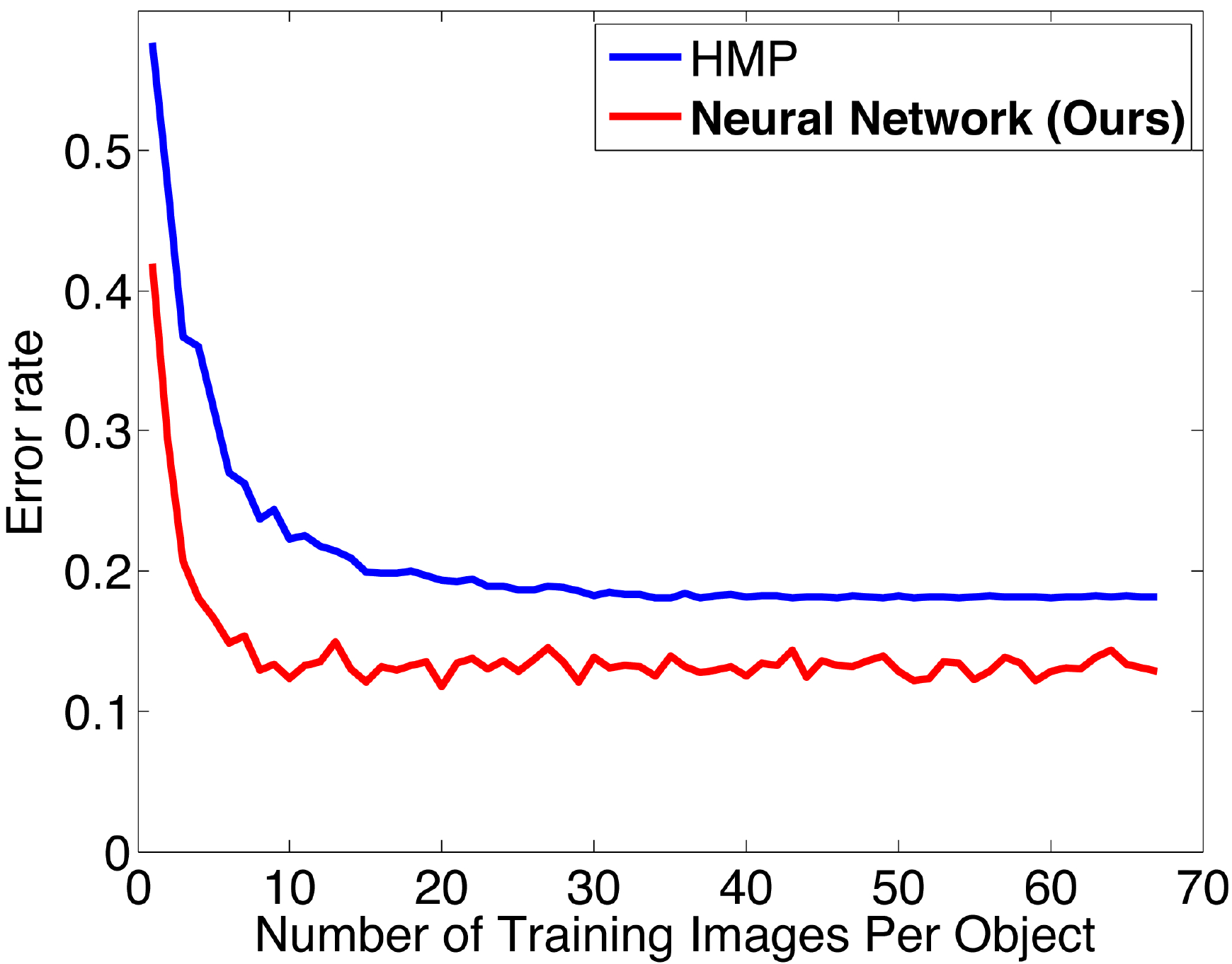}
\caption{We observe the effect on performance as we vary the number of training examples for a neural network as well as for the HMP baseline.  The neural network that we evaluate here is pre-trained only on ImageNet, with no multi-view pre-training.  The y-axis in this plot is the error rate (not accuracy).  These results are not directly comparable to those of Table~\ref{table:dense_training_tab} because in this setup we are training on only a 30 degree elevation angle, whereas for Table~\ref{table:dense_training_tab} we trained on both 30 and 60 degree elevation angles.  
}
\label{fig:sparse_training}
\end{figure}

Figure~\ref{fig:sparse_training} shows the performance as we vary the number of
views available during training.  We compare the performance of the best methods
of Section~\ref{sec:dense_training_setup}: HMP and a neural network pre-trained
only on ImageNet (without multi-view pre-training).  

As can be seen in this figure, the neural network  saturates performance after about 10 training images.  On the other hand, HMP~\cite{bo2013unsupervised} requires 30 training examples to saturate performance, and the result is still worse than that of the neural network.  
However, both methods perform poorly when only a single training example is available for each object.  Because this is a situation that occurs often in practice, we would like to focus our attention on this scenario, which we call ``one-shot learning for instance recognition."  We will show that, when we have only one training example per object, we can improve the performance of a neural network by performing multi-view pre-training.

\subsection{One-shot Learning for Instance Recognition}
\label{sec:one_shot}

\subsubsection{Baseline Methods}

\begin{table*}
\centering
\begin{tabular}{| l | >{\centering\arraybackslash} m{3cm} | >{\centering\arraybackslash} m{3cm} | >{\centering\arraybackslash} m{3cm} | }
\hline
& \multicolumn{3}{|c|}{\% Accuracy}  \\
\cline{2-4}
Method & Overall & Textured & Untextured \\
\hline
Random guessing & 0.3 & 0.3 & 0.3\\
BRISK \cite{leutenegger2011brisk}& 1.6 & 2.6 & 1.3 \\
ORB \cite{rublee2011orb} & 1.9 & 3.5 & 1.3\\
SURF \cite{bay2006surf} & 3.4 & 5.3 & 2.6\\
BOLD \cite{tombari2013bold} & 5.2 & 5.9 & 4.9 \\
SIFT \cite{lowe2004distinctive}& 6.3 & 12.6 & 3.9\\
Line-2D \cite{hinterstoisser2012gradient} & 5.5 & 0.3 & 7.4\\
Color Histogram Intersection \cite{swain1991color} & 12.4 & 23.3 & 8.2 \\
HMP \cite{bo2013unsupervised} & 42.3 & 53.8 & 37.9 \\
Neural Network (Ours) & 59.2 & 63.2 & 57.6 \\
\textbf{Neural Network, MV + BG pre-train (Ours)} & \textbf{63.9} & \textbf{73.8} & \textbf{60.0}\\
\hline
\end{tabular}
\caption{One-Shot Instance Recognition: We compare our neural network approach to previous methods when only a single view of each object is available during training.  The last row is our method with multi-view pre-training against random background images.}
\label{table:one_shot_training}
\end{table*}
In the next experimental setup, we are given only a single training example of each object.  At test time, we would like to recognize each object from novel viewpoints.  We  use the same test set as in Section~\ref{sec:dense_training_setup}, making this a strictly harder (though more realistic) training scenario.  For all objects, we train on only one training image at a 30 degree elevation angle, and we test on many different azimuthal viewpoints at a 45 degree elevation angle.

The results for this setup are shown in Table~\ref{table:one_shot_training}.  
The keypoint-matching based methods perform poorly, ranging from 1.6\% accuracy
for BRISK \cite{leutenegger2011brisk} to 6.3\% accuracy for
SIFT~\cite{lowe2004distinctive}.  More details
about the baseline methods, as well as a further analysis of their performance,
can be found in the appendix.

\begin{figure*}[htb]
\centering
\includegraphics[width=0.98\textwidth]{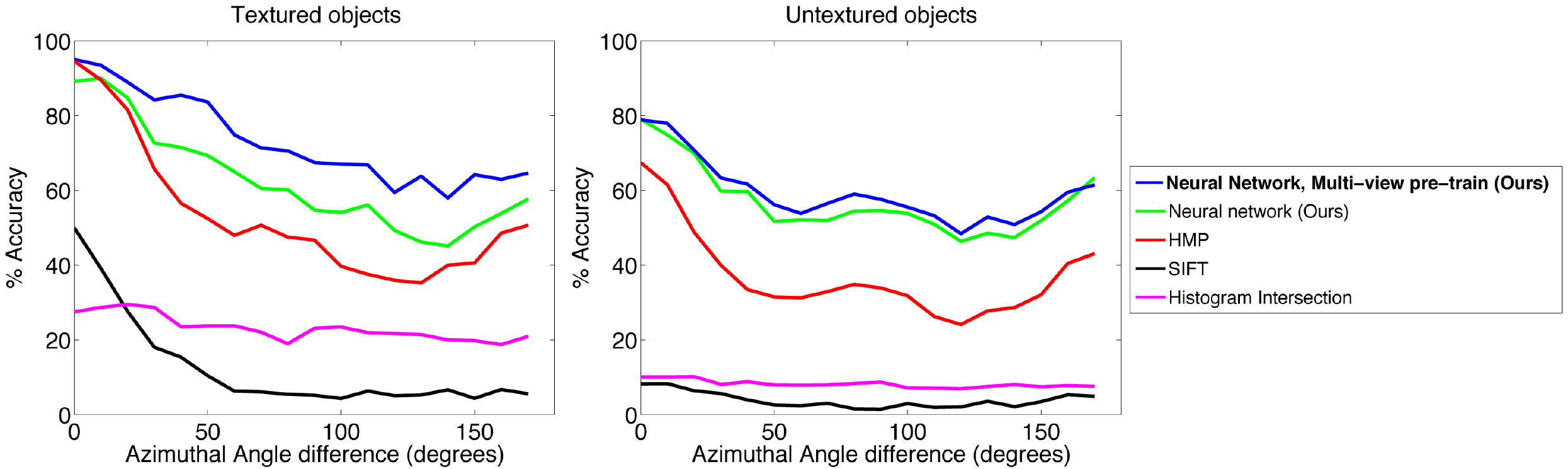}
\caption{Average accuracy as a function of the azimuthal angle difference between test examples and the corresponding training example.  Note that in all cases there is a 15 degree elevation difference between training and test images.  The machine learning methods have a small increase in performance near 180 degrees due to the rotational symmetry of some of the objects.}
\label{fig:angle_diff}
\end{figure*}

Machine learning methods perform significantly better on this task than the previous approaches.  This is surprising because these methods are trained on just a single example per object, which is not common for machine learning approaches.  HMP performs significantly better than the previous approaches when trained on just a single image, with an accuracy of 42.3\%.

As can be seen in Figure~\ref{fig:angle_diff}, HMP performs well when the test example is viewed from a similar angle as the training example.  However, the performance drops off quickly as the angular difference between the training and test example increases.  
Note that, although we are varying the azimuthal angle difference from 0 to 180 degrees, all of the images have an additional 15 degree elevation angle difference between training and test.
Given just a single training example, HMP is unable to find a good linear decision boundary that is viewpoint-invariant.  

\begin{figure}[htb]
\centering
\includegraphics[width=0.4\textwidth]{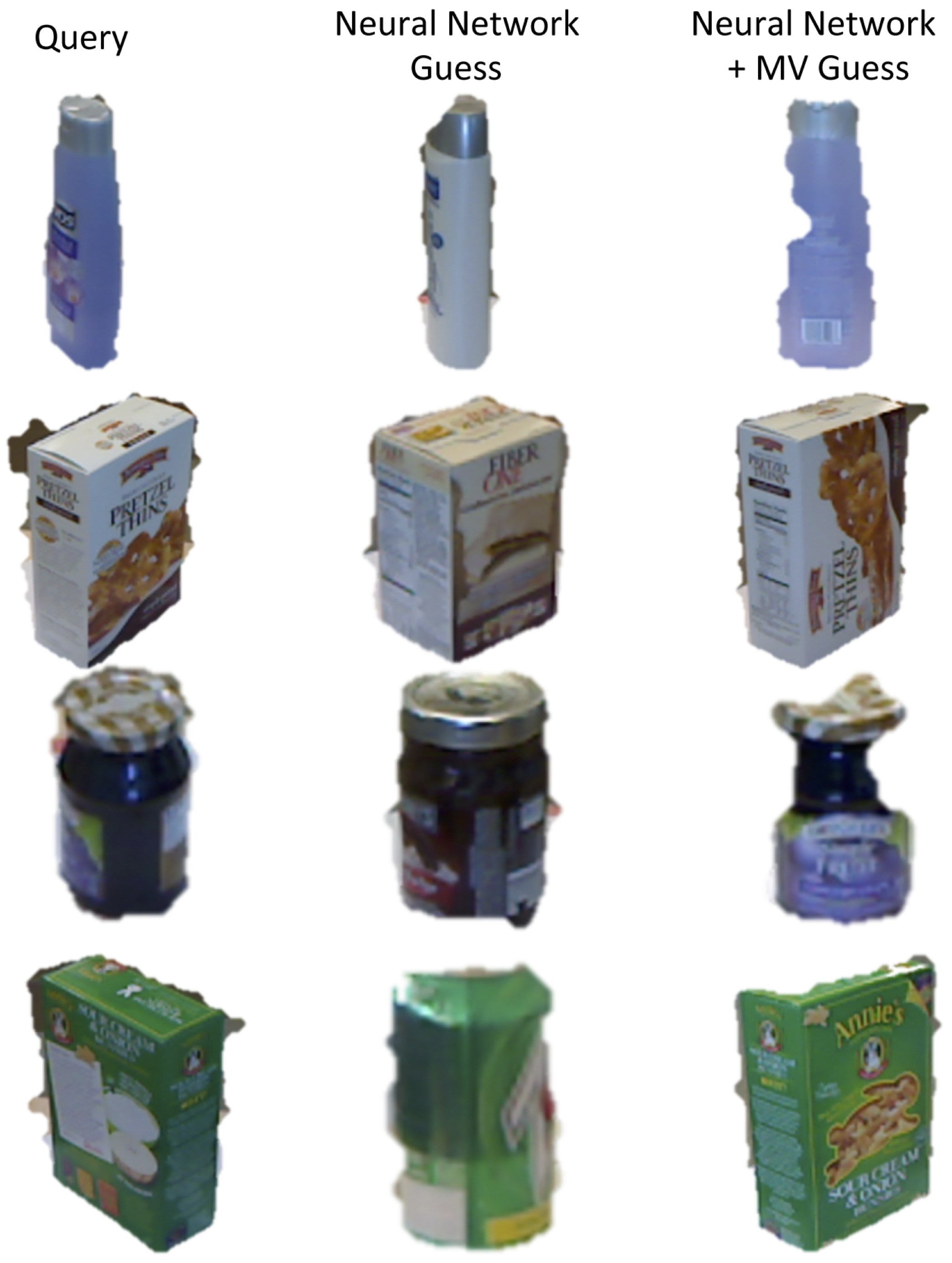}
\caption{Examples that were classified correctly using multi-view pre-training but were incorrectly classified using a neural network pre-trained only on ImageNet.  Left: Query image.  Middle: Guess by neural network pre-trained only on ImageNet (incorrect).  Right: Guess with multi-view pre-training (correct).}
\label{fig:errors}
\end{figure}

\subsubsection{Neural Networks}
We first evaluate the performance of a neural network that is pre-trained using ImageNet, as explained in section~\ref{sec:dense_training_setup}.  After pre-training on ImageNet and fine-tuning on our dataset with just a single image per object, our network achieves an accuracy of 59.2\%.  Compared to the next-best method, this is an absolute improvement in accuracy of 16.9\%, or a 29.3\% drop in the number of errors. More neural network baselines can be found in the appendix.

We next experiment to see if we can gain an additional benefit from incorporating a separate multi-view dataset via multi-view pre-training.  Note that the objects in the multi-view dataset are completely distinct from the 300 objects that we are trying to recognize.  For this experiment, we use the multi-view BigBird dataset~\cite{Singhetal_ICRA2014}.  This dataset consists of 125 objects recorded from many different viewpoints, and to ensure that we have no overlap with the set of test objects, we remove the box of White Cheddar Cheez-it crackers, which also appears in the RGB-D object dataset~\cite{lai2011large}.  We sample images from this dataset from 5 elevation angles and 20 azimuthal angles, for a total of 100 images per object.  The multi-view dataset that we incorporate thus consists of a total of 12,400 images from 124 objects.   

We evaluate the benefit of performing multi-view pre-training.  In this setup, after pre-training our network on the 1.2 million images from ImageNet, we further pre-train our network to perform 124-class classification with the 124 objects from the multi-view dataset.  Finally, we fine-tune the resulting network on the 300 objects from our single-view dataset, using just a single training example for each of the 300 objects.  The details of our training procedure are described in Section~\ref{sec:network_details}.

Multi-view pre-training is especially impactful at improving the recognition of textured objects.  By pre-training with a multi-view dataset, we obtain a 10.6\% absolute improvement (or a 28.8\% reduction in errors) on recognizing textured objects, compared to the neural network pre-trained only on ImageNet.  It is reasonable that multi-view pre-training gives a larger increase in performance on textured objects, since the appearance of textured objects changes more as a function of viewpoint compared to untextured objects.  Thus, training our network to be invariant to rotations gives an especially large benefit for recognizing textured objects from novel viewpoints.  At the same time, Table~\ref{table:one_shot_training} indicates that multi-view pre-training improves our performance for untextured objects as well.

Note that the multi-view dataset contains only 1\% as many images as were used in the original ImageNet pre-training step.  It is surprising that, given only 1\% more images, we obtain a 10.6\% improvement on the recognition of textured objects.  Figure~\ref{fig:errors} shows some examples of objects that our method was able to correctly recognize that were incorrectly recognized by a neural network pre-trained on ImageNet alone.  

\subsection{Objects in a Scene}
\label{sec:occlusion_handling}

In the previous set of experiments, we used test objects placed on a turntable so we could measure the rotational invariance of different methods in a controlled setting.  However, for most applications we would want to be able to detect objects in a full scene, with a real background and occlusions.  To measure whether our neural network with multi-view pre-training still gives the best performance in this more realistic setting, we used the RGB-D Scenes Dataset~\cite{lai2011large}.  This dataset has per-frame bounding box annotations, which makes it suitable for our evaluation purposes.  We crop the ground-truth bounding box from the scene and then classify the resulting image.  The results can be found in Table~\ref{table:one_shot_scene}.  As can be seen, multi-view pre-training improves performance even for objects placed in an indoor setting with real background and occlusions.  

\begin{table}
\centering
\begin{tabular}{| l | >{\centering\arraybackslash} m{3cm} | }
\hline
Method & \% Accuracy \\
\hline
Random guessing & 0.3\\
BRISK \cite{leutenegger2011brisk}& 9.4  \\
ORB \cite{rublee2011orb} & 6.6  \\
SURF \cite{bay2006surf} & 10.8\\
BOLD \cite{tombari2013bold} & 7.4 \\
SIFT \cite{lowe2004distinctive}& 12.9 \\
Line-2D \cite{hinterstoisser2012gradient} & 0.9\\
Color Hist Intersection \cite{swain1991color} & 9.2 \\
HMP \cite{bo2013unsupervised} & 25.4 \\
NN (Ours) & 41.0 \\
\textbf{NN + MV + BG (Ours)} & \textbf{44.1} \\
\hline
\end{tabular}
\caption{One-Shot Instance Recognition in a Scene: We train each method from just a single example and test on cropped images from a full scene, with occlusions and real backgrounds.  Note that the test set contains only a subset of the objects from Table~\ref{table:one_shot_training}, so the numbers are not directly comparable.}
\label{table:one_shot_scene}
\end{table}

Notice that the single-view objects from our training set were recorded while placed on a turntable.  To make our network robust to recognizing objects under novel backgrounds, the objects used for multi-view pre-training were synthetically placed against random scenes taken from the background category of the RGB-D scenes dataset~\cite{lai2011large}, as explained in Section~\ref{sec:joint_training}.  In Table~\ref{table:pre_train_bg_compare}, we demonstrate the advantage of this multi-view pre-training against a random background.  When the test images are depth-segmented, pre-training with a random background hurts performance slightly, with accuracy decreasing by 0.9\%.  However, when the test images are part of a real scene, pre-training with a random background increases robustness, improving accuracy by 2.6\%.  This demonstrates that pre-training on random backgrounds teaches our network to be robust to new backgrounds, even when the single-view objects being recognized are trained against a solid 
background.  

\begin{table}
\centering
\begin{tabular}{| l | >{\centering\arraybackslash} m{1.5cm} | >{\centering\arraybackslash} m{1.5cm} |  }
\hline
& \multicolumn{2}{|c|}{\% Accuracy}  \\
\cline{2-3}
Method & Segmented & In Scene\\
\hline
NN & 59.2 & 41.0 \\
\textbf{NN + MV (Ours)} & \textbf{64.8} & 41.5 \\
\textbf{NN + MV + BG (Ours)} & 63.9 & \textbf{44.1}\\
\hline
\end{tabular}
\caption{Comparison of different types of neural network pre-training.  All methods are initially pre-trained on ImageNet.  Top row: No multi-view pre-training.  Middle: Multi-view pre-training with a black background.  Bottom: Multi-view pre-training with random backgrounds.  Test images are either depth-segmented (left) or taken from a real scene with background included (right).}
\label{table:pre_train_bg_compare}
\end{table}

\begin{figure}[htb]
\centering
\includegraphics[width=0.23\textwidth]{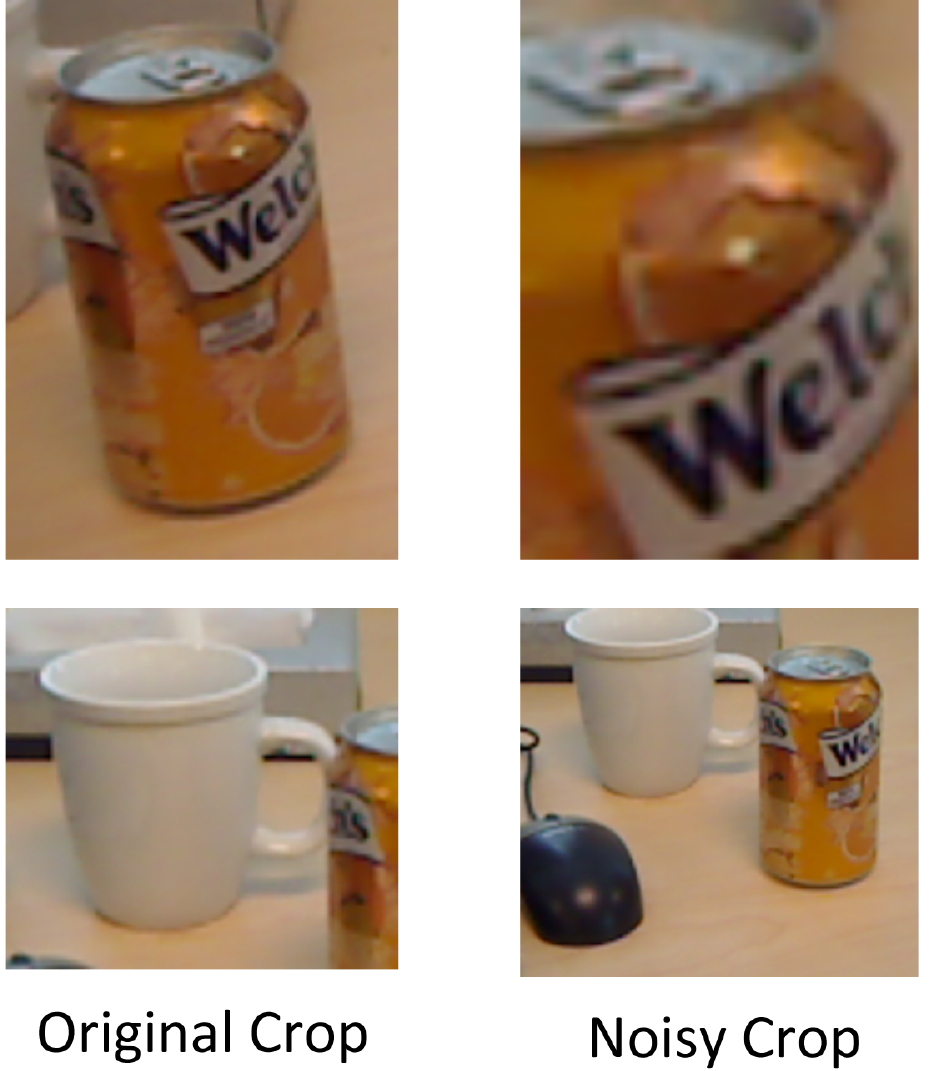}
\caption{Left: Crops from a scene used to test robustness to background and occlusions.  Right: The same crops with maximum noise added, to test robustness to bounding box noise.}
\label{fig:crops}
\end{figure}

\begin{figure}[htb]
\centering
\includegraphics[width=0.28\textwidth]{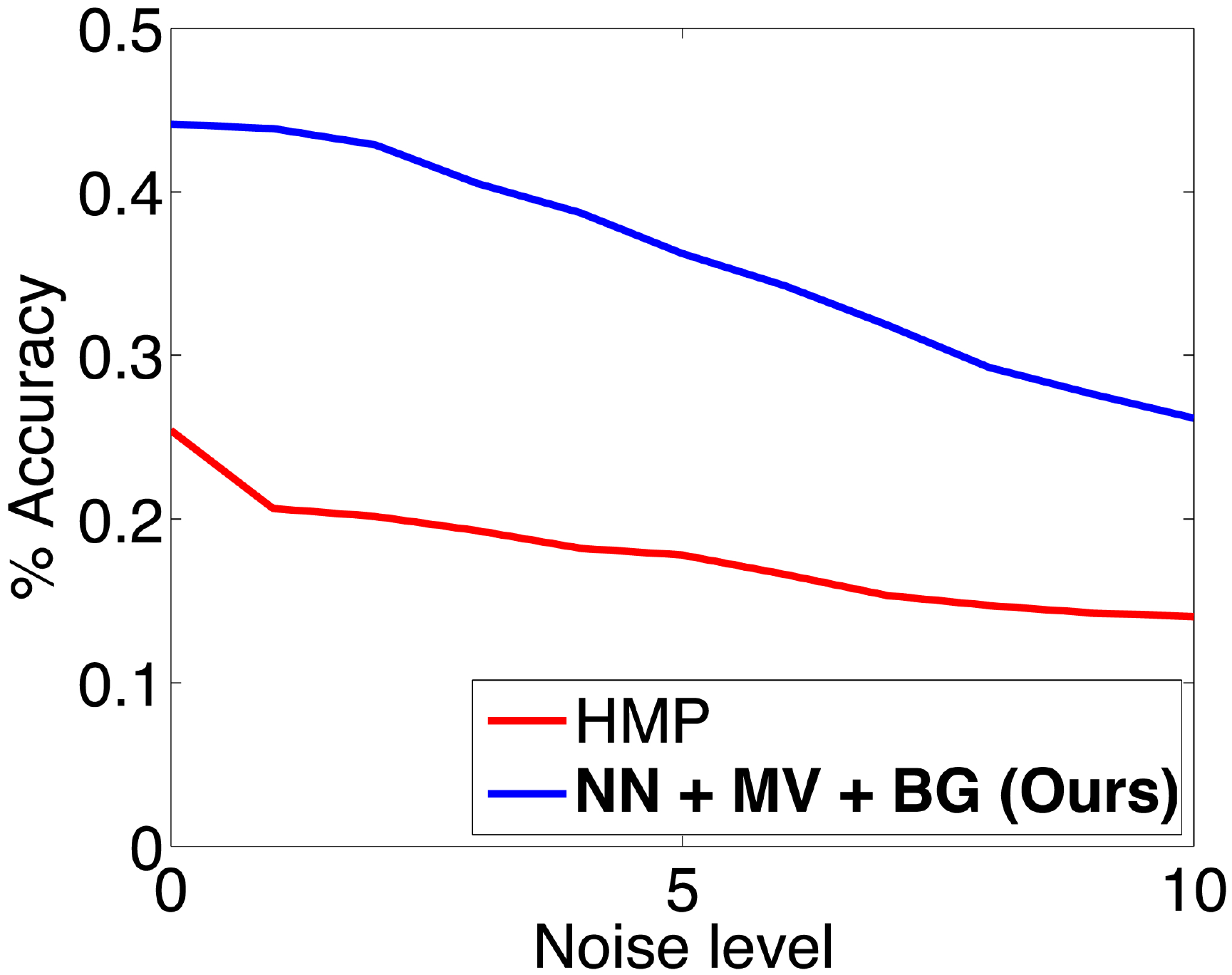}
\caption{Instance recognition accuracy as a function of the bounding box noise parameter n.}
\label{fig:noise_vs_accuracy}
\end{figure}

We also analyzed the performance as a function of the noise in the bounding box location.  To do this, for each bounding box we sampled a scaling factor $s$ and a displacement $\Delta x$ and $\Delta y$.  These values are sampled from a distribution that varies with a noise parameter $n$:
\begin{align}
s &\sim | \mathcal{N}(1, 0.025 n) | \\
\Delta x &\sim \mathcal{N}(0, 2 n) \\
\Delta y &\sim \mathcal{N}(0, 2 n)
\end{align}
The test crop locations are then scaled by the scaling factor and shifted by $\Delta x$ and $\Delta y$ pixels.
Examples of noisy images can be seen in Figure~\ref{fig:crops}.  Figure~\ref{fig:noise_vs_accuracy} shows the accuracy as a function of the noise parameter $n \in [0,10]$; as seen, our method is robust to noise in the bounding box location and still significantly outperforms the baseline methods.

\subsection{Multiview Pre-training analysis}
\label{sec:pre_training_analysis}
We can analyze which layers are benefiting most from multi-view pre-training.  Recall that, for our experiments, we hold the convolutional layers fixed, as determined by cross-validation using a hold-out validation set (Section~\ref{sec:network_details}).   
Table~\ref{table:pre_train_fix_layers} shows the effect of fixing different layers during multi-view pre-training, evaluated on the RGB-D objects dataset.  
If we hold the convolutional and both fully connected layers fixed, then we get the baseline performance (equivalent to not using multi-view pre-training).  If we fine-tune just the fc7 layer during multi-view pre-training, then we get an improvement  in performance of 1.7\%.  If we fine-tune both fc6 and fc7, then we get an additional improvement of 3\% (for a total benefit of 4.7\% over the baseline).  Finally, if we also fine-tune the convolutional layers, then we get a further improvement of 1.2\%.  Thus, the biggest improvement seems to come from fine-tuning fc6.  It seems that multi-view pre-training teaches the fully-connected layers the appropriate relationships between the convolutional features so that the network can be robust to viewpoint changes.

\begin{table}[htb]
\centering
\begin{tabular}{| l | >{\centering\arraybackslash} m{3cm} | }
\hline
Method & \% Accuracy \\
\hline
Baseline (no fine-tuning) & 59.2\\
Fine-tuning fc7 & 60.9 \\
Fine-tuning fc6 + fc7  & 63.9 \\
Fine-tuning all & 65.1\\
\hline
\end{tabular}
\caption{Classification accuracy when fixing different numbers of layers during multi-view pre-training.}
\label{table:pre_train_fix_layers}
\end{table}

\section{Conclusion}
We are able to train a neural network to recognize objects from novel viewpoints given only a single training image of each object. 
By pre-training our network with multiple views of a separate set of objects, the network learns an increased robustness to viewpoint changes compared to pre-training only on class-level datasets.  
We show that neural networks with multi-view pre-training outperform previous state-of-the-art methods for instance recognition on both textured and untextured objects.  

Thus, multi-view pre-training can make neural networks more robust to viewpoint changes.  
We also demonstrate that our multi-stage pre-training technique can be extended to learn other types of invariances, such as changes in background, by pre-training on objects with random backgrounds.  We hope to extend this approach to pre-train on tracked objects from videos in the wild, allowing our network to learn to be more robust to occlusions, lighting changes, and many other types of changes that an object can undergo in the real world.

{\scriptsize
\bibliographystyle{ieee}
\bibliography{iccv_instance_recognition_arxiv2}
}

\appendix
\section{Baseline Analysis}
\begin{figure*}[htb]
\centering
\includegraphics[width=0.98\textwidth]{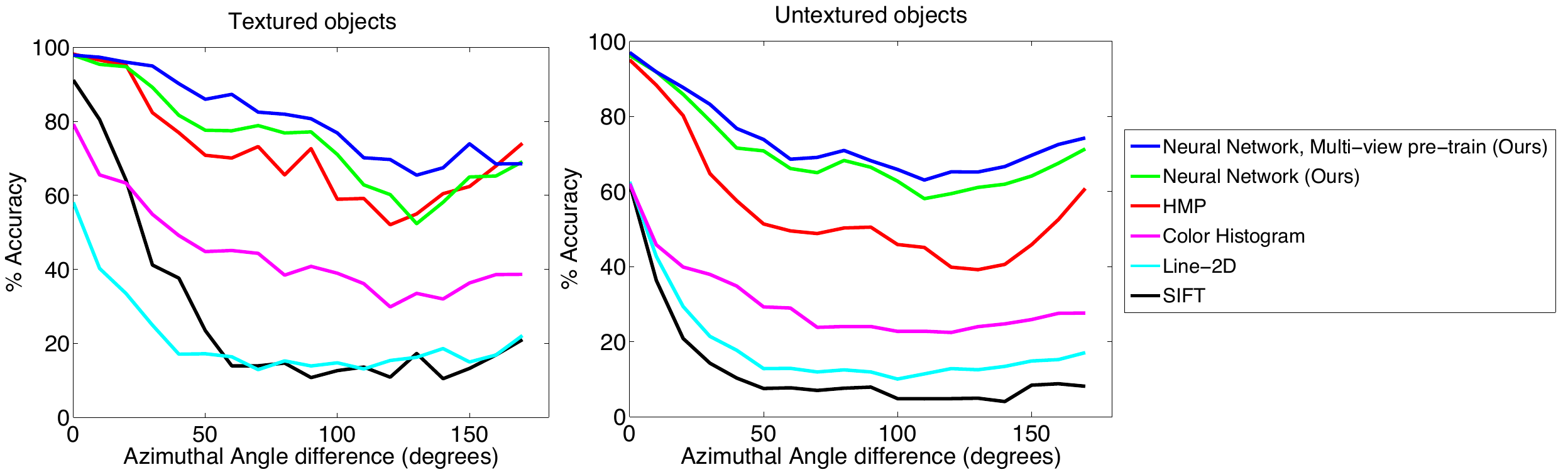}
\caption{Average accuracy as a function of the azimuthal angle difference between test examples and the corresponding training example. For this experiment, there is no elevation difference between training and test images (only an azimuthal difference). Some methods have a small increase in performance near 180 degrees due to the rotational symmetry of some of the objects.}
\label{fig:no_elevation_change}
\end{figure*}

In our previous analysis, test examples are viewed from a 15 degree elevation increase with respect to the training image, and performance is shown as the azimuthal angle is varied.  In Figure~\ref{fig:no_elevation_change}, we show the performance as the azimuthal angle is varied with no elevation difference between training and test images.  When the azimuthal angle difference is small, most methods have a very good performance, especially for textured objects.  However, for SIFT, Line-2D, and the color histogram, the performance quickly drops as the angle difference increases, showing a lack of robustness to changes in viewpoint.  The neural network with multi-view pre-training is the most robust to changes in viewpoint, as demonstrated by this figure.

\section{SIFT matching}
\label{sec:SIFT_RANSAC_COMPARE}
In our previous analysis, we compared the results of our neural-network based approach to that of SIFT-matching. 
Here, we compare SIFT-matching with and without a RANSAC geometric verification step.  We find that, for our dataset, we get lower overall accuracies when using SIFT with a geometric verification step, on both textured and untextured objects, as shown in Table~\ref{table:sift_ransac}.  Because this result is rather unintuitive, we now provide a detailed analysis explaining why adding a RANSAC geometric verification step can sometimes hurt performance. 

\begin{table*}[htb]
\centering
\begin{tabular}{| l | >{\centering\arraybackslash} m{3cm} | >{\centering\arraybackslash} m{3cm} | >{\centering\arraybackslash} m{3cm} | }
\hline
& \multicolumn{3}{|c|}{\% Accuracy}  \\
\cline{2-4}
Method & Overall & Textured & Untextured \\
\hline
SIFT \cite{lowe2004distinctive}& 6.3 & 12.6 & 3.9\\
SIFT with RANSAC geometric verification\cite{lowe2004distinctive}& 5.8 & 11.8 & 3.6\\
\hline
\end{tabular}
\caption{Performance of SIFT keypoint-matching, with and without using a RANSAC geometric verification step.  Overall, adding RANSAC hurts performance on this dataset, though it does help for some examples as well.  See Appendix Section~\ref{sec:SIFT_RANSAC_COMPARE} for further discussion.}
\label{table:sift_ransac}
\end{table*}

In both cases, we use a ratio test to filter matches, by comparing the distance from each SIFT match to that of its second-closest match, as was done in~\cite{lowe2004distinctive}.  We search over thresholds for the ratio test and choose the best threshold, separately with and without the geometric verification step.  We find that, when not using a geometric verification step, the highest accuracy is achieved by filtering with a ratio threshold of 0.6, whereas with a geometric verification step, the highest accuracy is achieved by filtering with a ratio threshold of 0.7.  Note that a ratio threshold of 1 (between the closest match and the second-closest match) would be equivalent to not using a ratio test for filtering at all.  Thus, for SIFT both with and without RANSAC geometric verification, filtering keypoints with a ratio test improves performance, as expected~\cite{lowe2004distinctive}.

\begin{figure*}[h!]
\centering
\includegraphics[width=0.98\textwidth]{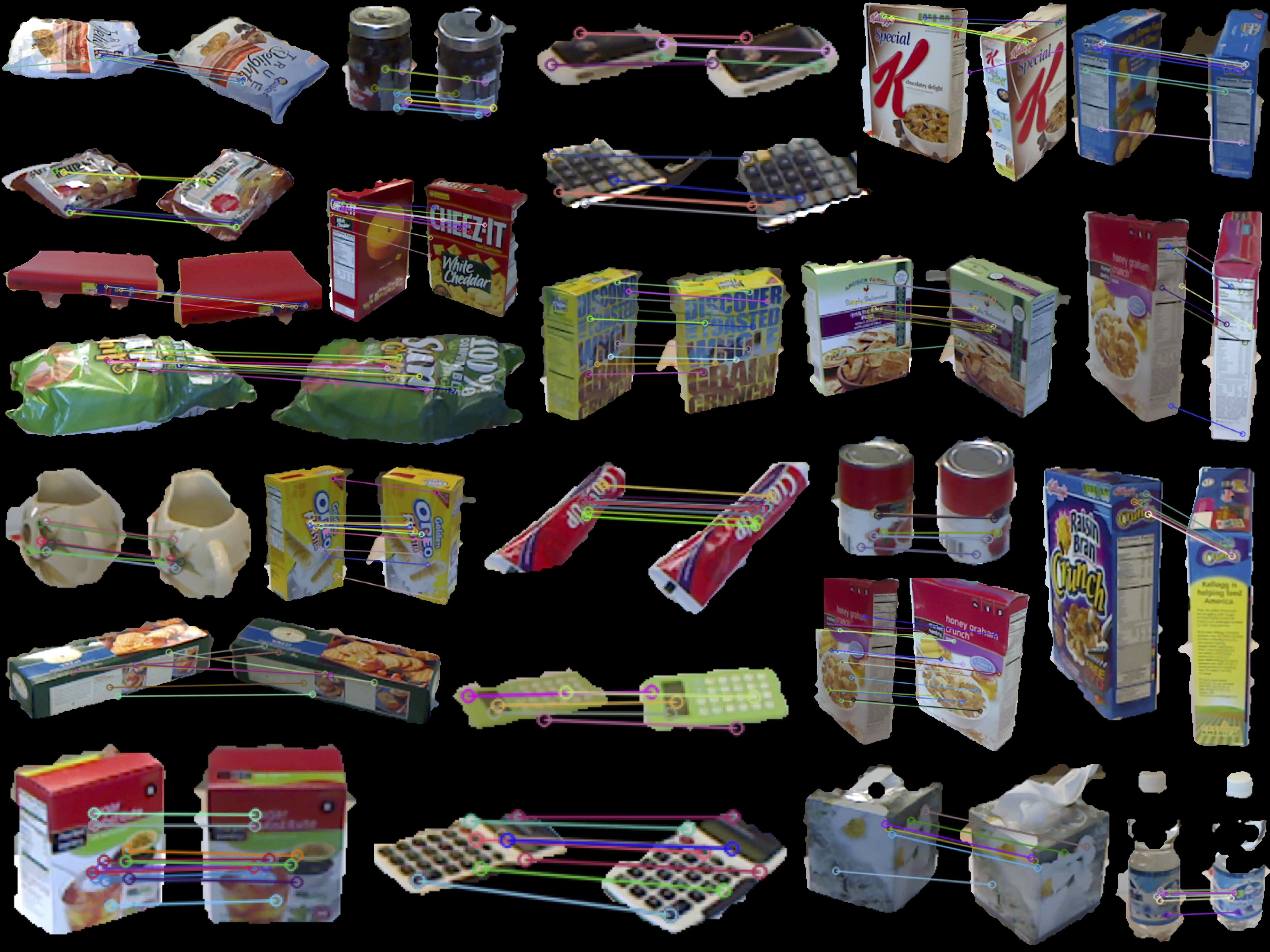}
\caption{Examples that were classified correctly using SIFT with a RANSAC geometric verification step but were classified incorrectly when SIFT was used without a geometric verification step.  On some objects, adding a geometric verification step can improve performance, as expected.}
\label{fig:ransac_improvements}
\end{figure*}

For some objects, adding a geometric verification step does improve SIFT performance.  We show in Figure~\ref{fig:ransac_improvements} a collection of examples for which SIFT with a geometric verification step is able to return the correct answer, whereas SIFT without geometric verification returns an incorrect answer.  These examples include highly textured objects with significant image structure, for which SIFT is able to match a logo or design on the object across viewpoints.

\begin{figure*}[htb]
\centering
\includegraphics[width=0.98\textwidth]{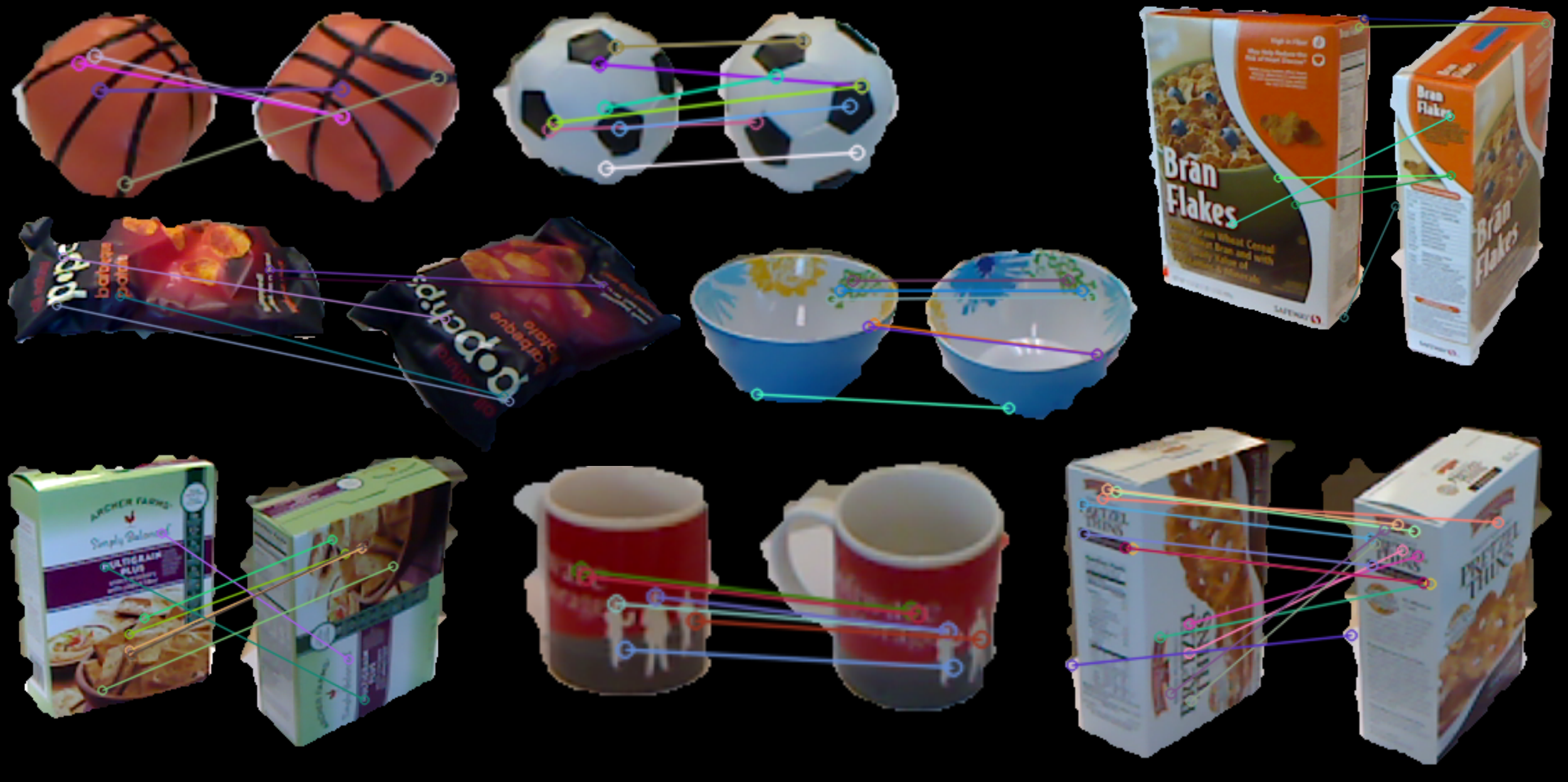}
\caption{Examples that were classified correctly when SIFT was used without a geometric verification step but were classified incorrectly when adding a RANSAC geometric verification step.  On some objects, adding a geometric verification step can actually hurt performance.  Note that the boxes on the bottom left and bottom right are attempts to match the front to the back of the packaging, which have similar elements placed in a different arrangement.}
\label{fig:ransac_errors}
\end{figure*}

However, in other cases, adding a geometric verification step can (surprisingly) hurt performance.  See Figure~\ref{fig:ransac_errors} for examples which were classified correctly using SIFT keypoint-matching but were classified incorrectly when a geometric verification step is added.  As can be seen in this figure, geometric verification can hurt performance when an object has repeated patterns.  For example, on the soccer ball, keypoints on the black pentagons in the training image are matched to different pentagons in the test image in a way that is not geometrically consistent.  When geometric verification is used, the soccer ball is not correctly matched because of the geometric inconsistency in these keypoint matches.  A similar effect is seen with the other items in Figure~\ref{fig:ransac_errors}, which have similar patterns that appear in different arrangements across the training and test images.  For such examples, SIFT gives better performance without a geometric verification step.  Still, the difference is relatively small, as shown in Table~\ref{table:sift_ransac}.

\section{Color Histogram}
For the color histogram method evaluated in the paper, we compute a histogram with 50 bins for hue and 60 bins for saturation.  We then compare the training and test image histograms using the histogram intersection kernel~\cite{swain1991color}.  We also compare histograms using a correlation, chi-squared distance, and Bhattacharyya distance, and found that the histogram intersection kernel performed the best.

\section{Neural Network Baselines}
We compare our multi-view pre-training procedure to a few alternative neural network training procedures, and show that our method significantly outperforms the alternatives.  First, one might be concerned that, by fine-tuning the network on such a small dataset of 300 objects with just a single image per object, we would overfit our network.  We therefore compare to a procedure in which we hold the entire network fixed and fine-tune only the final layer. 

 We also try using the output of different layers as an input to a linear SVM or a nearest-neighbor lookup, as has previously been done~\cite{razavian2014cnn, zeiler2014visualizing}.  For the nearest-neighbor lookup, because we have just 1 training example per class, we use kNN with k = 1.  For the linear SVM, we use a one-vs-one SVM with voting~\cite{razavian2014cnn}.  Because each class has just a single training example, changing the C-value does not significantly affect performance; for our experiments we use C = 1.

The results are shown in Table~\ref{table:fine_tune_final}.  The three approaches that just use features from a fixed network pre-trained on ImageNet perform poorly.  The nearest-neighbor approach performs the worst, with an accuracy of 49\% using fc7 features, and the SVM and softmax approaches perform only slightly better.  Fine-tuning the entire network improves performance by 6.6\% compared to training just the top layer.  This is surprising because we are fine-tuning with just 300 training images, with only 1 image per class.  As shown in the paper, a neural network with multi-view pre-training performs the best, improving performance by 4.7\% overall and by 10.6\% for textured objects.

\begin{table*}[htb]
\centering
\begin{tabular}{| l | >{\centering\arraybackslash} m{3cm} | }
\hline
Method & \% Accuracy \\
\hline
kNN with pool5 \cite{razavian2014cnn} & 34.3  \\
kNN with fc6 \cite{razavian2014cnn} & 47.6 \\
kNN with fc7 \cite{razavian2014cnn} & 49.0  \\
\hline
SVM with pool5 \cite{razavian2014cnn, zeiler2014visualizing} & 39.2 \\
SVM with fc6 \cite{razavian2014cnn, zeiler2014visualizing} & 50.7  \\
SVM with fc7 \cite{razavian2014cnn, zeiler2014visualizing}& 50.2 \\
\hline
Neural Network, Fine-tune only top layer & 52.6  \\
Neural Network, Fine-tune all  & 59.2  \\
\textbf{Neural Network, MV + BG pre-train, Fine-tune all (Ours)} & \textbf{63.9} \\
\hline
\end{tabular}
\caption{Comparison of our method, in which we fine-tune the entire neural network (bottom), to approaches which hold all of the weights fixed except for the final layer.}
\label{table:fine_tune_final}
\end{table*}

\section{Multiview Pre-training Analysis}
\subsection{Comparing Pre-training Strategies}
Given that the original network was trained with 1 million images from ImageNet, it is surprising that we get a 4.7\% overall improvement, and a 10.6\% improvement on textured objects, after multi-view pre-training with just 1\% more images.  The number of images from the multi-view dataset is relatively small, and the objects in this dataset are distinct from the objects in our training and test sets.  These results suggest that multi-view pre-training teaches our network to be robust to changes in viewpoint in a way that pre-training with images taken from separate object instances (\eg ImageNet) does not.

However, one might still ask whether the improvement in performance is really due to the multi-view pre-training procedure or whether it is simply the result of our network seeing an additional 1\% more images.  Although this seems unlikely, we evaluate this hypothesis by comparing the performance of two training setups.  

In both experiments, we start by pre-training our network on ImageNet.  
We then pre-train our network in two different ways.
In the first multi-view pre-training approach, we pre-train our network using 20 views of each multi-view object, for a total of 2,480 training images (``Multi-view pre-train").  In the second pre-training approach, we have a separate class for each of 20 poses and each of 124 object instances, for a total of 2,480 classes  (``Pose-class pre-train").  We use the same 2,480 training images in both cases.  After pre-training our network using each method, we then fine-tune the network on the single-view object instances that we wish to recognize.  

In summary, in the first setup, we have one class per object and we are training from multiple views of that object; in the second setup, we have one class per pose per object.  However, in both cases, the training images being used are exactly the same; the only difference is the way in which we have defined the classes for multi-view pre-training.  By comparing the performance of the network using these two different intermediate training procedures, we can determine whether the benefit of multi-view pre-training comes from learning viewpoint invariance due to our training procedure, or whether the benefit is simply the result of learning from more total images.  Because the total number of images is the same in both cases, we can directly analyze the effect of our multi-view pre-training procedure.

\begin{table*}
\centering
\begin{tabular}{| l | >{\centering\arraybackslash} m{3cm} | }
\hline
Method & \% Accuracy \\
\hline
Neural Network & 59.2 \\% & 63.2 & 57.6 \\
Neural Network, Pose-class pre-train, 2,480 images & 40.9 \\ %
\textbf{Neural Network, Multi-view pre-train, 2,480 images (Ours)} & \textbf{62.3} \\ %
\hline
\end{tabular}
\caption{We compare two versions of pre-training: one with a separate class per object and per pose (middle row) and one with just a separate class per object (bottom row).  Both methods use the exact same images, but by using them in a different way we see a difference in the effect on performance.  The top row is a neural network without any multi-view pre-training.  All networks are initially pre-trained using ImageNet.}
\label{table:multi-view-analysis}
\end{table*}

The results are shown in Table~\ref{table:multi-view-analysis}.  Using the standard neural network (pre-trained on ImageNet), we get 59.2\% accuracy.  If we use multi-view pre-training on 2,480 images, we get 62.3\% accuracy.  On the other hand, if we pre-train a separate class for each pose of each object using the same 2,480 images, accuracy drops to 40.9\%.  

 In the first case, our network learns to be viewpoint invariant and shows improved performance on the final single-view dataset.  In the second case, our network learns to distinguish between poses, and shows a significant drop in performance on the final single-view dataset.  In the second case, even though the network was pre-trained on the same set of images, they were not used in a way that allowed the network to learn to be viewpoint invariant.  Thus, the improvement of multi-view pre-training is not just that the network has ``seen" more images, but by pre-training to classify objects from different viewpoints, the network learns to be robust to changes in viewpoint.

\subsection{Comparing to Data Augmentation}
We propose to use multi-view pre-training as a way to teach our network to be robust to changes in viewpoint.  However, another strategy to train a network to be robust to changes in viewpoint is to simulate changes in viewpoint via data augmentation.  From just a single image, this is difficult for non-planar objects.  Still, we can simulate perspective warps, which will be a correct transformation for fronto-parallel planar objects, and might still be a useful transformation for non-planar objects.

The results are shown in Table~\ref{table:data-augmentation}.  Using perspective warps during data augmentation does increase the robustness of the network to new viewpoints.  To the best of our knowledge, this is a novel type of data augmentation that has not been explored previously.

\begin{table*}
\centering
\begin{tabular}{| l | >{\centering\arraybackslash} m{3cm} | }
\hline
Method & \% Accuracy \\
\hline
Neural Network & 59.2 \\% & 63.2 & 57.6 \\
Neural Network, Perspective Augmentation & 62.9 \\ %
\textbf{Neural Network, MV + BG pre-train (Ours)} & \textbf{63.9} \\ %
\hline
\end{tabular}
\caption{We compare multi-view pre-training (bottom) to data augmentation by perspective warping (middle).  The top row is a neural network without any multi-view pre-training or perspective warping.  All networks are initially pre-trained using ImageNet.}
\label{table:data-augmentation}
\end{table*}

Using multi-view pre-training, our method gives an even bigger improvement.  However, besides the effect on performance, there are a number of other reasons to perform multi-view pre-training rather than (or in addition to) performing data augmentation.  First, our results are obtained from multi-view pre-training with just 124 objects.  If more objects are used for this stage, then our performance might be expected to improve even further.  Additionally, we demonstrate that multi-view pre-training can teach our network to be robust to both viewpoint changes as well as changes in background.  By tracking objects that undergo other types of changes, our network can use multi-stage pre-training to learn to be robust to even more types of changes, such as changes in lighting, occlusions, or deformations.  Multi-stage pre-training is thus a general technique that can be used to learn different invariances by observing how objects change their appearance over time.  

\begin{figure}[htb]
\centering
\includegraphics[width=0.48\textwidth]{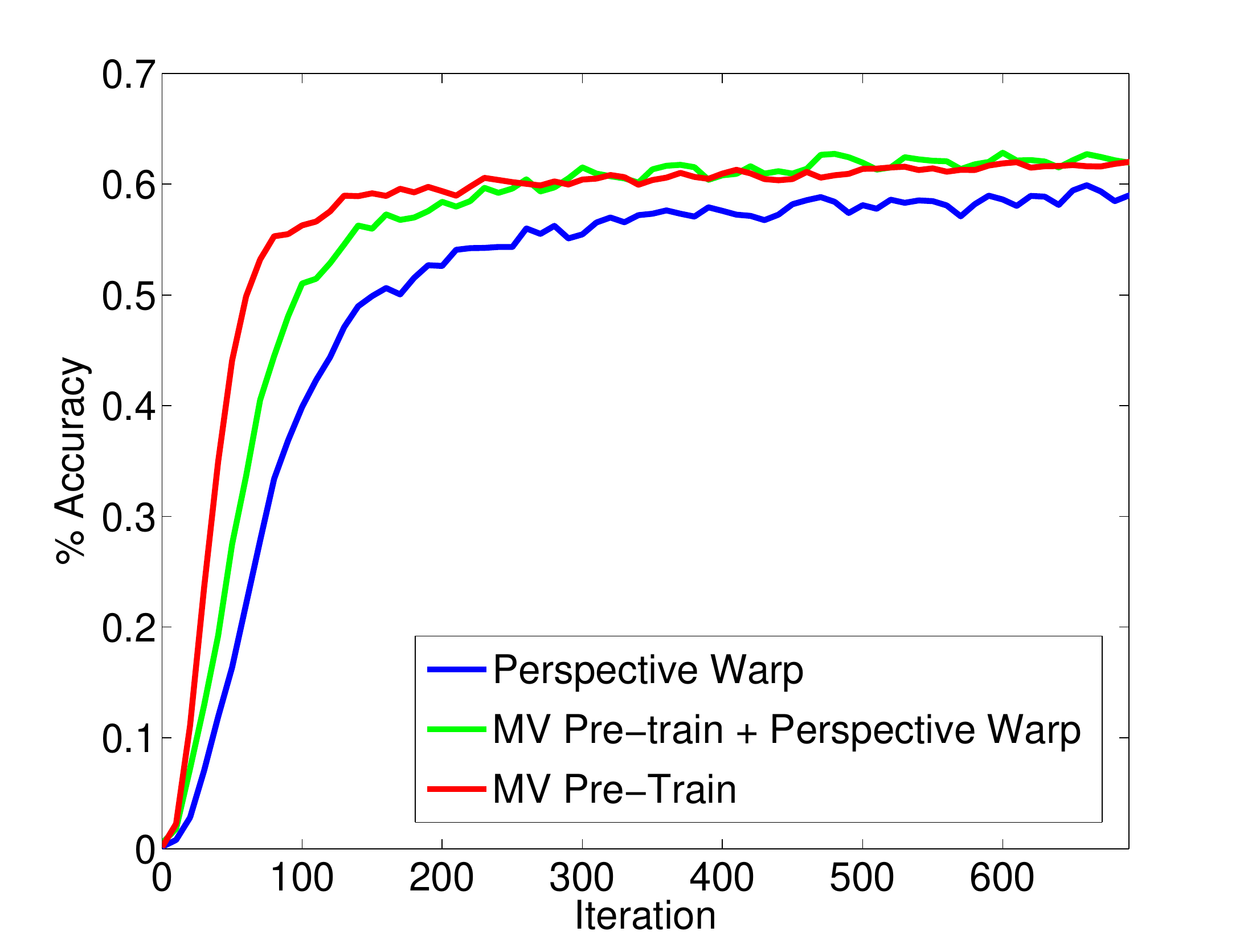}
\caption{Convergence rates for three different approaches to learning robustness to changes in viewpoint.  Multi-view pre-training allows our network to quickly learn to recognize new objects.}
\label{fig:convergence}
\end{figure}

Finally, multi-view pre-training leads to a faster convergence, as shown in Figure~\ref{fig:convergence}.  Because the method has already learned to be robust to changes in viewpoint, it can quickly learn to recognize new objects from novel viewpoints, without the need for extensive data augmentation.  The figure shows that the slowest convergence is achieved using data augmentation with perspective warps.  Faster convergence is achieved if multi-view pre-training was used so that the network has already learned to be robust to changes in viewpoint.

\section{Error Analysis}
\begin{figure}[htb]
\centering
\includegraphics[width=0.48\textwidth]{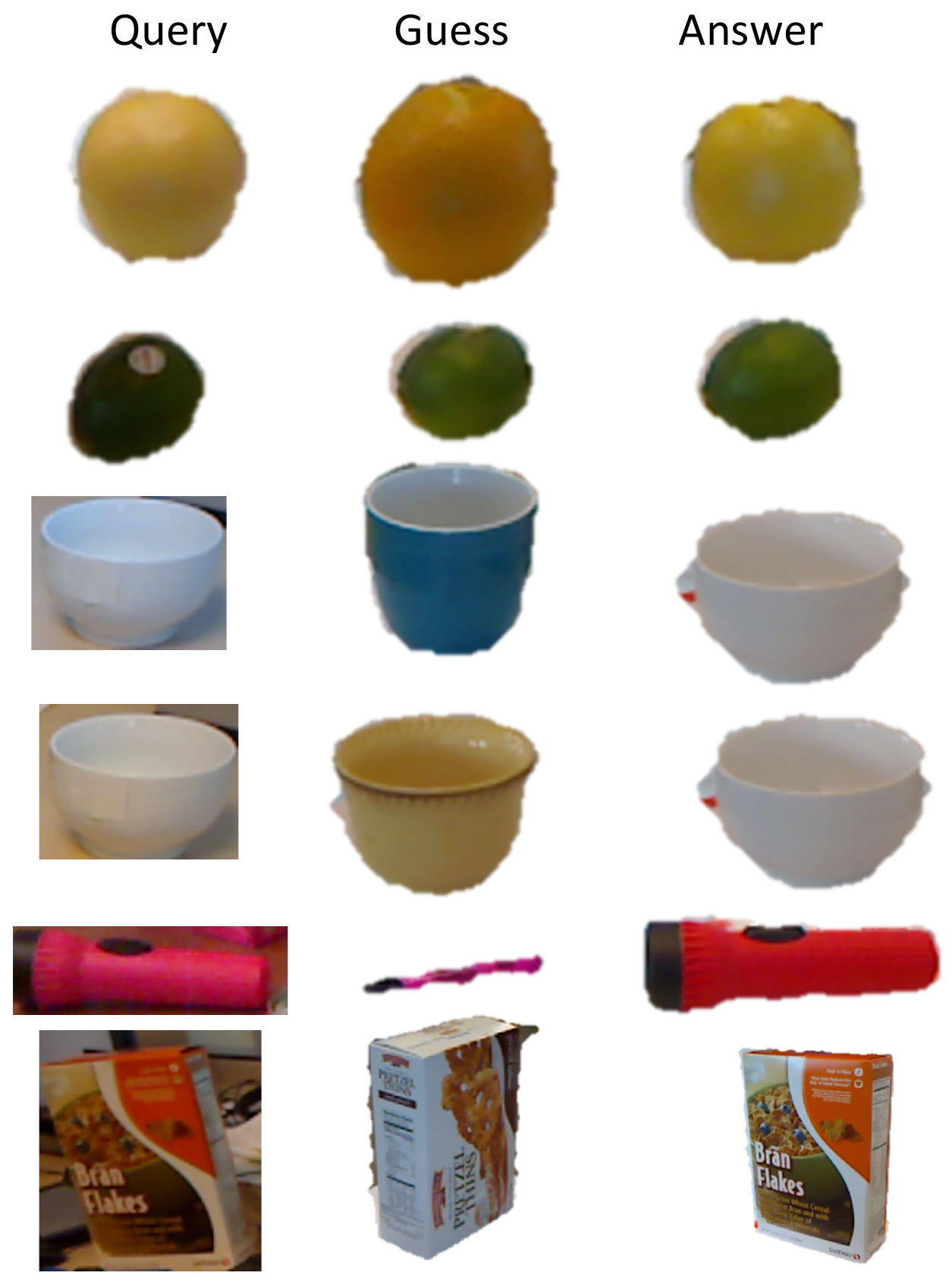}
\caption{Examples that were classified incorrectly even after performing multi-view
pre-training. These examples demonstrate that our network is not robust to subtle color differences (top two rows) or changes in lighting (bottom 4 rows).
Left: query image. Middle:
guess by neural network with multi-view pre-training.
Right: the correct match.}
\label{fig:color_changes}
\end{figure}

\begin{figure}[htb]
\centering
\includegraphics[width=0.48\textwidth]{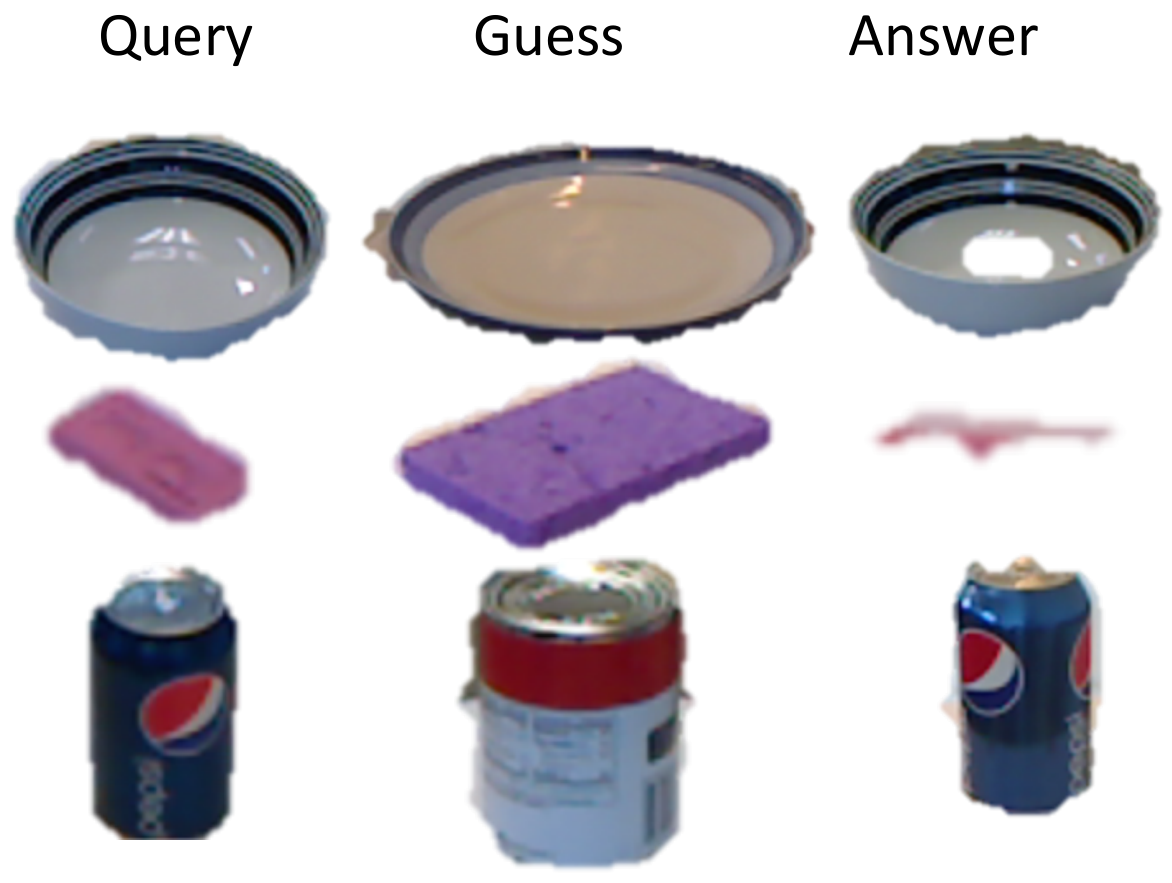}
\caption{Examples that were classified incorrectly even after performing multi-view
pre-training. These errors seem to be caused by the depth-segmentation that is used to pre-process the images~\cite{lai2011large}.
Left: query image. Middle:
guess by neural network with multi-view pre-training.
Right: the correct match.}
\label{fig:sensor_errors}
\end{figure}

\begin{figure}[htb]
\centering
\includegraphics[width=0.48\textwidth]{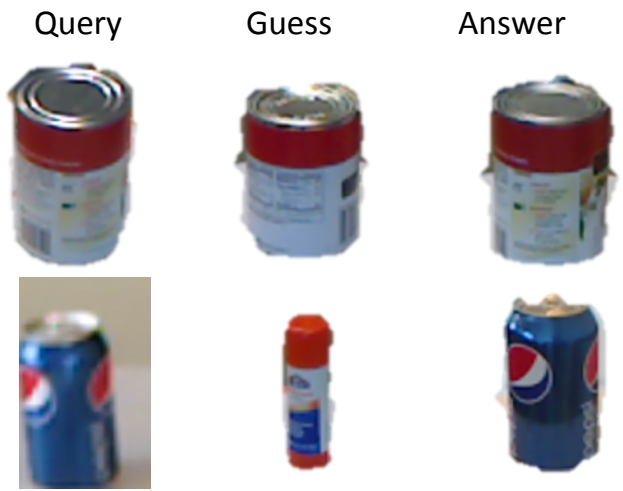}
\caption{Examples that were classified incorrectly even after performing multi-view
pre-training. Some errors are caused by the limited viewpoint available during training (top) or poor bounding box localization for objects in a scene (bottom).
Left: query image. Middle:
guess by neural network with multi-view pre-training.
Right: the correct match.}
\label{fig:other_errors}
\end{figure}

It is instructive to look at what kind of errors our network still makes after multi-view pre-training.  Although multi-view pre-training teaches our network to be robust to changes in viewpoint, there are other types of changes that our network is still not robust to.  For example, Figure~\ref{fig:color_changes} shows that our network is not robust to subtle color differences between objects, or changes in color due to the scene lighting.  Some errors are caused by the depth-segmentation that is used to pre-process the images (Figure~\ref{fig:sensor_errors}).  We use the pre-processed images from Lai, et al~\cite{lai2011large}.  Other errors are caused by the limited viewpoint available during training, or by poor bounding-box localization of objects in a scene, as shown in Figure~\ref{fig:other_errors}.

\section{Experimental Details}
\begin{figure}[htb]
\centering
\includegraphics[width=0.48\textwidth]{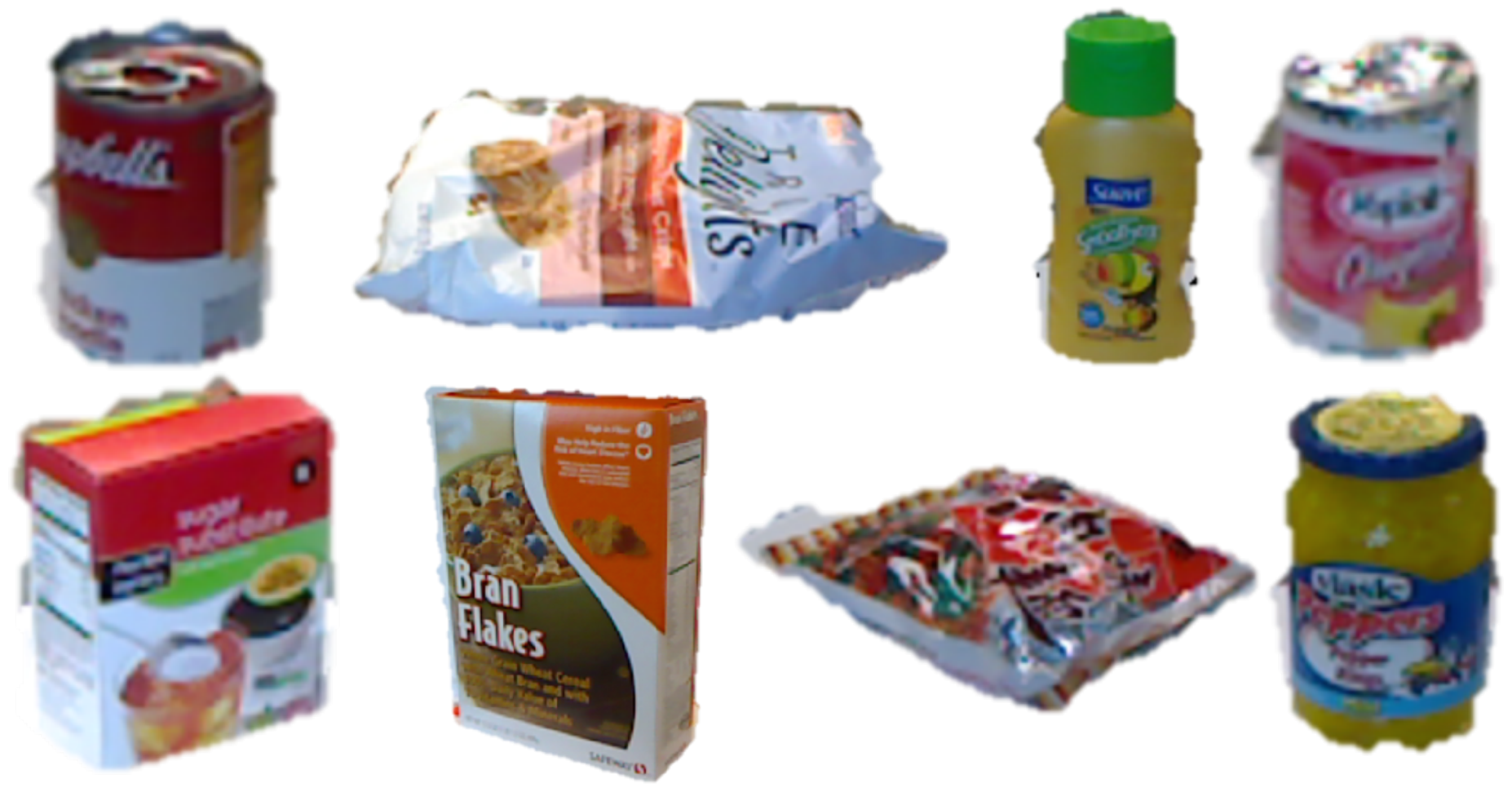}
\caption{Examples of textured objects from the RGB-D object dataset.}
\label{fig:textured}
\end{figure}

\begin{figure}[htb]
\centering
\includegraphics[width=0.48\textwidth]{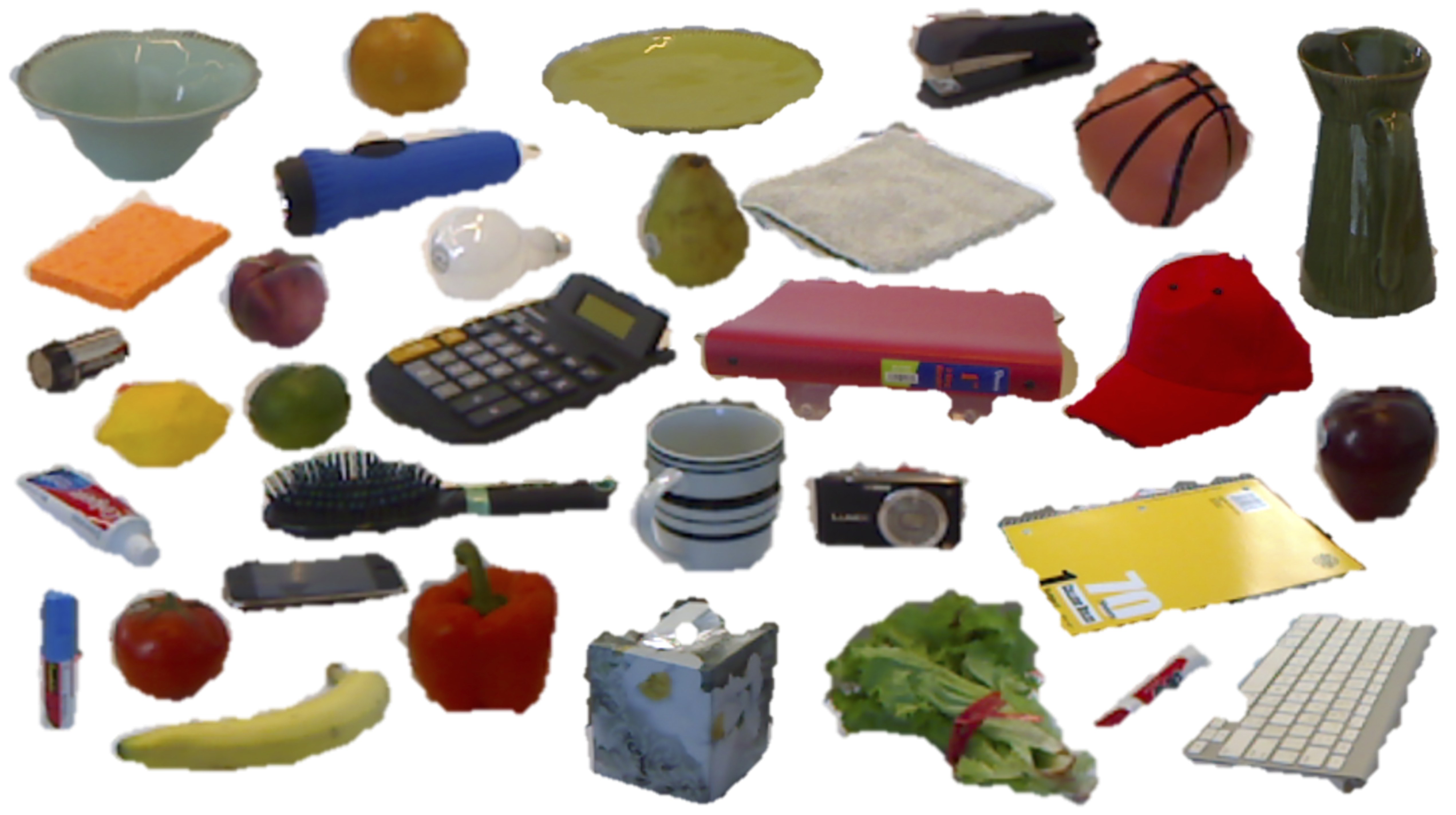}
\caption{Examples of untextured objects from the RGB-D object dataset.}
\label{fig:untextured}
\end{figure}

In many of our experiments, we compute performance separately for textured vs untextured objects.  These divisions were chosen by the authors, and we considered the following object categories to be textured: cereal box, food bag, food box, food can, food cup, food jar, instant noodles, shampoo, soda can, and water bottle (see Figure~\ref{fig:textured}).  The remaining categories were considered to be untextured: apple, ball, banana, bell pepper, binder, bowl, calculator, camera, cap, cell phone, coffee mug, comb, dry battery, flashlight, garlic, glue stick, greens, hand towel, keyboard, kleenex, lemon, light bulb, lime, marker, mushroom, notebook, onion, orange, peach, pear, pitcher, plate, pliers, potato, rubber eraser, scissors, sponge, stapler, tomato, toothbrush, and toothpaste (see Figure~\ref{fig:untextured}).  Some objects could have been placed into either category, but the main purpose is to observe general trends in recognizing textured vs untextured objects, so individual categorization choices are less important.

\end{document}